\documentclass[11pt,a4paper]{article}
\usepackage{authblk}
\usepackage[hyperref]{acl2019}
\usepackage{times}
\usepackage{latexsym}
\usepackage{graphicx}
\usepackage{url}
\usepackage{amsmath}
\usepackage{amssymb}
\usepackage{multirow}
\usepackage{tablefootnote}
\usepackage{multicol}

\aclfinalcopy % Uncomment this line for the final submission
\def\aclpaperid{***} %  Enter the acl Paper ID here

\title{Best Practices for Learning Domain-Specific Cross-Lingual Embeddings}

\author{%
Lena Shakurova\textsuperscript{1, 2}, Beata Nyari\textsuperscript{1}, Chao Li\textsuperscript{1}, Mihai Rotaru\textsuperscript{1} \\
1 Textkernel B.V., Amsterdam, Netherlands \\
2 Radboud University, Nijmegen, Netherlands \\
\footnotesize \texttt{\{shakurova,nyari,chaoli,rotaru\}@textkernel.nl}}

\date{}

\begin{document}
\maketitle
\begin{abstract}
Cross-lingual embeddings aim to represent words in multiple languages in a shared vector space by capturing semantic similarities across languages. They are a crucial component for scaling tasks to multiple languages by transferring knowledge from languages with rich resources to low-resource languages. A common approach to learning cross-lingual embeddings is to train monolingual embeddings separately for each language and learn a linear projection from the monolingual spaces into a shared space, where the mapping relies on a small seed dictionary. While there are high-quality generic seed dictionaries and pre-trained cross-lingual embeddings available for many language pairs, there is little research on how they perform on specialised tasks. In this paper, we investigate the best practices for constructing the seed dictionary for a specific domain. We evaluate the embeddings on the sequence labelling task of Curriculum Vitae parsing and show that the size of a bilingual dictionary, the frequency of the dictionary words in the domain corpora and the source of data (task-specific vs generic) influence the performance. We also show that the less training data is available in the low-resource language, the more the construction of the bilingual dictionary matters, and demonstrate that some of the choices are crucial in the zero-shot transfer learning case. 
\end{abstract}

\section{Introduction}

Expanding Natural Language Processing (NLP) models to new languages typically involves creating completely new data sets for each language which comes with challenges such as acquiring and annotating the data. To avoid these tedious and costly tasks, one can use cross-lingual embeddings to enable knowledge transfer from languages with sufficient training data to low-resource languages. 

Cross-lingual embeddings aim to represent words in multiple languages in a shared vector space by capturing semantic similarities across languages. Based on the assumption that the embedding spaces of different languages exhibit a similar structure \cite{Mikolov:2013b}, previous work proposed to learn a linear transformation which projects independently learned monolingual spaces into a single shared space, using a seed translation dictionary \cite{Faruqui:2014}.
Although more advanced techniques involving jointly optimising monolingual and cross-lingual objectives were proposed, most of these solutions require some form of cross-lingual supervision via parallel data \cite{guo-etal-2015-cross, klementiev-etal-2012-inducing, xiao-guo-2014-distributed, Hermann2014MultilingualDR, sogaard-etal-2015-inverted, DBLP:journals/corr/VulicM15}. However, for applications targeting a specific domain (in our case, human resources) there is often little to no parallel data available, so simple alignment-based methods relying on only a small translation dictionary remain an attractive choice.

We adopt the Multilingual CCA framework \cite{Ammar:2016}, and evaluate the cross-lingual embedding on a sequence labelling task in Curriculum Vitae parsing domain. We use this framework as it only requires an easier to acquire seed dictionary. Previous work has shown that the quality of this dictionary influences the cross-lingual embeddings \cite{vulic-korhonen-2016-role}. However, to the best of our knowledge, there has been no extensive research on the choice of a seed dictionary in a non-generic domain. In addition, little attention was paid to how the quality of the bilingual dictionary affects performance as some labelled data from the target language is added.

In this paper, we investigate the best practices to create a seed dictionary for training domain-specific cross-lingual embeddings. We measure the impact of different choices of the dictionary creation on the downstream task: the dictionary size, the source of the words and their frequency, in both zero-shot and joint training scenarios.

\section{Related work}

\textbf{Offline linear map induction methods} \label{CCA} The earliest approach to induce a linear mapping from the monolingual embedding spaces into a shared space was introduced in \cite{Mikolov:2013a}. They propose to learn the mapping by optimising the least squares objective on the monolingual embedding matrices corresponding to translational equivalent pairs. Subsequent research aimed to improve the mapping quality by optimising different objectives such as max-margin \cite{Lazaridou:2015} and by introducing an orthogonality constraint to the bilingual map to enforce self-consistency \cite{Xing:2015,Smith:2017}. \cite{Artetxe:2016} provide a theoretical analysis to existing approaches and in a follow-up research \cite{Artetxe:2018} they propose to learn principled bilingual mappings via a series of linear transformations.

An extensive survey of different approaches, including offline and online methods can be found in \cite{DBLP:journals/corr/Ruder17}.

\textbf{The role of bilingual dictionary} A common way to select a bilingual dictionary is by using either automatic translations of frequent words or word alignments. For instance, \cite{Faruqui:2014} select the target word to which the source word is most frequently aligned in parallel corpora. \cite{Mikolov:2013b} use the 5,000 most frequent words from the source language with their translations. To investigate the impact of the dictionary on the embedding quality, \cite{vulic-korhonen-2016-role} evaluate different factors and conclude that carefully selecting highly reliable symmetric translation pairs improves the performance of benchmark word-translation tasks. The authors also demonstrate that increasing the lexicon size over 10,000 pairs show a slow and steady decrease in performance.

\section{Task} 
In this work, we look at the Curriculum Vitae (CV) parsing task: extraction of relevant information (e.g. name, job titles, etc) from a given CV and converting it into a structured format. This task can be cast as a cascaded sequence labelling problem \cite{Yu:2005} consisting of two steps: section segmentation and extraction of pre-defined entities, similar to named entity recognition task (NER). In the first step, a model segments the entire CV into sections such as personal information, education, experience or skills. In the second step, for each section, a dedicated model extracts entities specific to that section such as name, phone number, etc. from personal section and degree level, institution, etc. from education section. For all models, we use the standard BIO approach (Begin, Inside, Outside) to sequence labelling \cite{DBLP:journals/corr/cmp-lg-9505040}. For brevity, in this paper, we present the results of extracting 2 entities from the experience section: job title and organisation name.

\section{Methodology}

We conduct the experiments for German-English and Dutch-English cross-lingual embeddings. Given a bilingual seed dictionary, we use the learned CCA linear projection (see Section \ref{CCA}) between the monolingual vector spaces to project German/Dutch embeddings into the English space. The projected embeddings are then fed into the sequence labelling model. 
The sequence labelling model is always trained in the English space using either English training data (zero-shot) or English training data combined with projected German/Dutch training data. The model is tested using projected German/Dutch embeddings and German/Dutch test data.
We experiment with several factors in the construction of the bilingual dictionary: source of data, size, and the frequency of the bilingual dictionary entries in the domain corpus.

\subsection{Training data} 
\textbf{Monolingual embeddings} For each language, we train monolingual word2vec embeddings \cite{word2vec} on normalised CV data. The dimension of embeddings is 150, vocabulary size is 169k, 503k and 286k for English, German and Dutch respectively (minimum frequency 5).

\textbf{Corpora} In our experiments, we use English as high resource language and German and Dutch as low resource. The number of annotated documents is 4342 for English, 1947 for German and 2383 for Dutch. Having enough resources for German/Dutch also allows us to study the impact of increasing the amount of training data. Each document contains on average 11 entities. We split our data into train, development and test set with proportions of 70, 15 and 15\% accordingly. 

\subsection{Bilingual dictionary factors}

\textbf{Source of data (IDP vs MUSE vs domain)}: We want to investigate the impact of constructing the bilingual dictionary from domain-specific words versus employing generic seed dictionaries: 1) from Facebook's MUSE project\footnote{https://github.com/facebookresearch/MUSE} 2) from The Internet Dictionary Project (IDP)\footnote{http://www.june29.com/IDP/}. MUSE dictionaries were specifically created for developing cross-lingual embeddings \cite{lample2017unsupervised}, whereas IDP dictionaries were produced for the purpose of making royalty-free translating dictionaries accessible to the Internet community. For the domain-specific dictionary, we picked top frequent words (see below) from the source monolingual corpus (German/Dutch) and translated the selected words into English using Yandex Translate API\footnote{https://pypi.org/project/yandex-translater/}. Stop words were removed and the words shorter than three characters were filtered out due to their unreliable translation.

\textbf{Frequency of bilingual dictionary entries (high vs lower)}:  We compared choosing most frequent words to those selected from a lower frequency range (between top 5-10\%) in our domain-specific corpus. It has been observed by previous research that due to the fact that frequent terms are over-represented in commonly used seed dictionaries, the performance of cross-lingual mappings is much lower on rare words \cite{Nakashole:2018}. Motivated by this finding we wanted to analyse the downstream effect of adding rarer terms to the dictionary.

\textbf{Size of bilingual dictionary (1k vs 5k vs 10k)}: We compared seed dictionaries of different size: 1.000, 5.000 and 10.000. Understanding the impact of this factor is important as larger dictionaries are more expensive to create. 

\textbf{Validation}: Previous research suggests using back-translation as a verification step for a translation pair. We skipped this because we noticed that certain words are crucial to be included in the seed dictionary and despite their translation being correct often they would be invalidated because of synonyms or suffixes (e.g. \textit{pers\"{o}nliche} $\rightarrow$ \textit{personal} $\rightarrow$ \textit{pers\"{o}nlich}). Instead, we filter words whose translations do not reach a frequency threshold in the English corpus, where this threshold is tuned on a validation set.

\subsection{Model Architecture} 
Our sequence labelling model is a stacked Bidirectional LSTM with a CRF layer based on \cite{huang2015bidirectional} with a pre-trained embedding layer. We used Adam optimiser and trained for 150 epochs. The network's hyperparameters are tuned on the English development set.

\subsection{Evaluation metrics} 
As extrinsic evaluation metric of the cross-lingual embeddings, we use the average F1 score across the 2 entities we extract (job title and organisation name). As intrinsic evaluation metric, we use the precision at 1 (P@1) measured on the MUSE test sets consisting of 1,500 translation pairs.

\begin{table*}[t]
\resizebox{\textwidth}{!}{%
\begin{tabular}{lllllll}
\multicolumn{1}{c}{\multirow{2}{*}{\textbf{Factor combinations}}} & \multicolumn{3}{c}{\textbf{DE - EN}}                                             & \multicolumn{3}{c}{\textbf{NL-EN}}      \\
\multicolumn{1}{c}{}                                              & \multicolumn{1}{c}{Joint training} & \multicolumn{1}{c}{Zero shot} & P@1 & Joint training & Zero shot & P@1 \\ \hline
\textbf{IDP} + 5k               & 79.5                               & 61.5                          & 1.1         &  -             &  -         &   -          \\
\textbf{MUSE} + 5k              & 80.4                               & 72.1                          & 0.8         & 81.4          & 77.2      & 2.1         \\
\textbf{domain} + 5k + high freq                 & 81.1                               & 75.8                          & 1.7         & 81.5          & 79.1      & 2.5         \\ \hline
domain + 5k + \textbf{high freq}                     & 81.1                               & 75.8                          & 1.7         & 81.4          & 79.1      & 2.5         \\
domain + 5k + \textbf{lower freq}                    & 81.0                               & 70.2                          & 1.0         & 80.9      & 71.6      & 1.9         \\ \hline
domain + \textbf{10k} + high freq                    & 81.5                               & 76.8                          & 1.2         & 81.6          & 79.3      & 2.8         \\
domain + \textbf{5k} + high freq                     & 81.1                               & 75.8                          & 1.7         & 81.4          & 79.1      & 2.5         \\
domain + \textbf{1k} + high freq                     & 80.1                               & 72.2                          & 1.2         & 79.3          & 77.8      & 1.6         \\ \hline
\end{tabular}%
}
\caption{Average F1 and precision@1 score for bilingual dictionary experiments. Joint training uses 200 documents from the low resource language.}
\label{tab:bilingual-dict}
\end{table*}

\begin{table*}[t!]
\resizebox{\textwidth}{!}{%
\begin{tabular}{lllll}
\multicolumn{1}{c}{\multirow{2}{*}{\textbf{Low resource data}}} & \multicolumn{2}{c}{\textbf{DE - EN}}                               & \multicolumn{2}{c}{\textbf{NL - EN}} \\
\multicolumn{1}{c}{}                                            & \multicolumn{1}{c}{Monolingual} & \multicolumn{1}{c}{Cross-lingual gain} & Monolingual   & Cross-lingual gain   \\ \hline
None (zero-shot)                                                & -                            & +75.8                               & -             & +79.1                \\
200 CVs                                                         & 77.0                           & +4.1                                & 75.1          & +6.3                 \\
500 CVs                                                         & 83.9                         & +0.2                                & 80.9          & +1.3                 \\
Full set                                                         & 87.1                         & +0.0                                  & 83.5          & +0.3                  
\end{tabular}%
}
\caption{Gain from knowledge transfer, averaged F1 score.  Full set is 1363 for German and 1678 for Dutch.}
\label{tab:zero-shot}
\end{table*}

\section{Results and discussion}

Table \ref{tab:bilingual-dict} presents our results on how the 3 bilingual dictionary factors influence the downstream task performance and the precision@1 score. We start with the best practices from previous work (top 5k frequent words) and change one factor at a time choosing the best performing setting when moving to the next factor.

From the first set of rows, we see that using in-domain seed words improves the task performance over generic dictionaries. This effect is amplified in the zero-shot transfer learning scenario. We also see that using a bilingual dictionary (MUSE) employed by previous NLP research performs much better than typical free online resource dictionary (IDP). These observations are particularly important in industry settings where it is a common practice to use free open-source resources. We also see that the intrinsic metric (P@1) yields very low scores and it is uncorrelated with the task metric e.g. it ranks MUSE and IDP in the reverse order. This highlights the importance of verifying cross-lingual embeddings on the downstream task.

We also observe that choosing less frequent seed words degrades the performance in the zero-shot case. Qualitative analysis shows that including certain high-frequency words can be crucial for our task: these words are typically section header words (e.g. \textit{Pers\"{o}nliche Angaben} (Personal Information)) or common context words of the entities of interest (e.g. \textit{Erfahrung} (experience)). Since these words tend to occur in similar contexts as the entities, they tend to be confused with these entities in the zero-shot setting if they are not in the dictionary. Being common words, their meaning is quickly picked up when jointly training with some German/Dutch data.

In terms of vocabulary size, we notice that even with a smaller 1k domain-specific dictionary we tend to get a competitive performance. Using 5k terms seems sufficient, although in line with \cite{vulic-korhonen-2016-role} we observe that a larger vocabulary (10k) gives only a slight improvement.

% Another interesting result is that all factors play a much bigger role in the zero-shot transfer learning case than in the joint training case. By analysing neighbourhoods of non-seed German words projected in the English space, we noticed that even though the nearest English neighbours are related words (e.g. job title words), often the distances are quite big. Our intuition is that, specifically for sequence labelling tasks, adding some training data from the low-resource language allows the BLSTM model to the learn about these nearby neighbourhoods and account for the leeway created by imperfect cross-lingual projections.

By analysing neighbourhoods of non-seed German words projected in the English space, we noticed that even though the nearest English neighbours are related words (e.g. job title words), often the distances are quite big. Our intuition is that, specifically for sequence labelling tasks, adding some training data from the low-resource language allows the BLSTM model to the learn about these nearby neighbourhoods and account for the leeway created by imperfect cross-lingual projections.

We investigate the impact of increasing the size of low-resource language data in Table \ref{tab:zero-shot}. For these experiments, we use the best performing seed dictionary (5k high-frequency words from domain corpus). The results demonstrate that with a strong English-only CV parsing model and cross-lingual embeddings we achieve comparable results to a model trained on only 15\% of the low-resource language. We also observe that the gain of transfer learning diminishes as we jointly train with an increasing amount of German data.

\section{Conclusions and future work}

In this paper, we investigate the best practices for constructing a bilingual dictionary for learning domain-specific cross-lingual embeddings. We show that for our CV parsing task, the dictionary should be created from top frequency domain-specific words. A dictionary size of 5k tends to be sufficient, with limited gains coming from adding more words. We also show that the less training data is available in the low-resource language, the more these best practices matter.

In future work, we plan to extend our research to cover other language pairs (e.g. Slavic languages) or more distant pairs (e.g. English-Russian). We also plan to look at cross-lingual subwords embeddings which become crucial for languages with more complex morphology. 

% \section*{Acknowledgments}

% The acknowledgments should go immediately before the references.  Do
% not number the acknowledgments section. Do not include this section
% when submitting your paper for review. \\

\end{document}